\documentclass[10pt]{wlscirep}
\usepackage{amsthm}
\usepackage{bm}
\usepackage{graphicx}
\usepackage{xcolor}
\usepackage{float}
\usepackage{hyperref}
\hypersetup{hidelinks}
\usepackage{caption}
\usepackage{mdframed}
\usepackage{fvextra} 
\usepackage{pifont}
\usepackage{booktabs,multirow,makecell,tabularx}
\usepackage{subcaption}
\usepackage{booktabs}
\usepackage{multirow}
\usepackage{rotating}
\usepackage{threeparttable}
\usepackage{array}
\usepackage{algorithmicx}
\usepackage[ruled,linesnumbered]{algorithm2e}

\title{Multimodal Benchmark for Safety Assessment in Industrial Inspection Scenarios}

\author[1,2]{Zeyi Liu}
\author[1,2,3]{Shuang Liu}
\author[3]{Jihai Min}
\author[3]{Zhaoheng Zhang}
\author[4]{Jun Cen}
\author[1,2]{Pengyu Han}
\author[1,2]{Songqiao Hu}
\author[3]{Zihan Meng}
\author[1,2,*]{Xiao He}
\author[1,5]{Donghua Zhou}
\affil[1]{Department of Automation, Tsinghua University, Beijing 100084, China}
\affil[2]{Institute for Embodied Intelligence and Robotics, Tsinghua University, Beijing 100084, China}
\affil[3]{TetraBOT Intelligence Co., Ltd., Nanjing 210000, China}
\affil[4]{DAMO Academy, Alibaba Group, Hangzhou 311100, China}
\affil[5]{School of Automation, Southeast University, Nanjing 210096, China}

\affil[*]{Corresponding author}

\begin{abstract}
With the rapid development of industrial intelligence and unmanned inspection, reliable perception and safety assessment for AI systems in complex and dynamic industrial sites has become a key bottleneck for deploying predictive maintenance and autonomous inspection. Most public datasets remain limited by simulated data sources, single-modality sensing, or the absence of fine-grained object-level annotations, which prevents robust scene understanding and multimodal safety reasoning for industrial foundation models. To address these limitations, InspecSafe-V1 is released as the first multimodal benchmark dataset for industrial inspection safety assessment that is collected from routine operations of real inspection robots in real-world environments. InspecSafe-V1 covers five representative industrial scenarios, including tunnels, power facilities, sintering equipment, oil and gas petrochemical plants, and coal conveyor trestles. The dataset is constructed from 41 wheeled and rail-mounted inspection robots operating at 2,239 valid inspection sites, yielding 5,013 inspection instances. For each instance, pixel-level segmentation annotations are provided for key objects in visible-spectrum images. In addition, a semantic scene description and a corresponding safety level label are provided according to practical inspection tasks. Seven synchronized sensing modalities are further included, including infrared video, audio, depth point clouds, radar point clouds, gas measurements, temperature, and humidity, to support multimodal anomaly recognition, cross-modal fusion, and comprehensive safety assessment in industrial environments.
\end{abstract}
\begin{document}
\flushbottom
\maketitle

\thispagestyle{empty}
\section*{Background \& Summary}

Industrial inspection environments are often characterized by high noise, severe occlusion, large illumination changes, specular reflections, and complex equipment layouts. High-risk events may also occur, including open flames or smoke, unauthorized human intrusion, equipment overheating, and abnormal ambient gas conditions \cite{thomas1995real,10375819,weimer2016design}. Since inspection robots are required to operate autonomously for long periods in such environments, AI systems with scene understanding and safety reasoning capabilities are urgently needed to enable a shift from post-event detection to early warning \cite{he2025realtime,gao2020real,10750900}.

In practical deployment, the main bottleneck of industrial AI is often not the model architecture, but the lack of high-quality data that can systematically capture the complexity and risk patterns of real industrial settings. High-quality data is a prerequisite for industrial safety intelligence \cite{khandakar2025harnessing,liu2023real,yu2019inspection,tavakoli2010development
}. Substantial domain shifts are common in industrial sites, driven by differences in operating conditions and equipment, seasonal and day-night illumination changes, specular reflections, dust and smoke occlusion, and sensor noise and drift \cite{liu2023dynamic,9960812,10458267}. When training data are collected under idealized conditions, limited viewpoints, or static scenes, model performance can degrade or even fail during deployment. Therefore, datasets that include real disturbances and robot-centric viewpoints can reduce the distribution gap between training and deployment and improve cross-scene robustness \cite{soori2024intelligent,mei2024replanvlm,zeng2023large,wang2025multi}.

Safety assessment in industrial environments also depends on more than object recognition \cite{hu2025vlsa,lind2008safety,le2013outlines,groenewald2014yarrowia,hao2022hazard}. Judgments are required on object states and their interactions, such as a person entering a hazardous area, trends of equipment overheating, and the coupling between abnormal gas levels and ventilation conditions. Learning such structured knowledge typically relies on fine-grained object-level annotations and synchronized alignment across multiple sensing modalities. Without these elements, reliable interpretable reasoning is difficult to achieve \cite{wozniak2025multi}. In addition, most industrial scenarios involve high risk, and algorithmic progress should be supported by unified evaluation protocols and reproducible benchmarks.  Without a standardized benchmark, fair comparison and iterative improvement become difficult, which further hinders engineering validation and large-scale deployment.

\begin{table*}[t]
\centering
\scriptsize
\renewcommand{\arraystretch}{1.15}
\setlength{\tabcolsep}{4pt}

\caption{Comparison with typical publicly available multimodal datasets.}
\label{tab:dataset_comparison}
\begin{tabularx}{\textwidth}{l l l l l l}
\toprule
\textbf{Dataset} &
\textbf{Domain} &
\textbf{Platform} &
\textbf{Scenes} &
\textbf{Modalities} \\
\midrule

\multicolumn{6}{l}{\textit{Industrial inspection / safety / anomaly}}\\
\midrule
\textbf{InspecSafe-V1 (Ours)} &
Safety Assessment &
Wheeled \& rail-mounted robots &
5 industrial scenarios &
\makecell[l]{RGB, TIR, Language, Audio, Depth, Radar, Gas, Temp, Hum} \\

MVTec AD &
Defect Detection &
Static rig (lab-like) &
15 object types &
RGB \\

VisA &
Defect Detection &
Static rig (lab-like) &
12 object types &
RGB \\

Real-IAD &
Defect Detection &
Static/multi-view setup &
30 products (5 views) &
RGB \\

MVTec 3D-AD &
Defect Detection &
3D scanner/rig &
10 object types &
RGB + 3D point cloud \\

\midrule
\multicolumn{6}{l}{\textit{Autonomous driving (multisensor)}}\\
\midrule
KITTI &
Driving &
Vehicle &
Urban/highway &
Stereo RGB, LiDAR, Radar (GPS/IMU) \\

nuScenes &
Driving &
Vehicle &
1,000 scenes &
RGB, LiDAR, Radar (GPS/IMU) \\

Waymo Open &
Driving &
Vehicle &
2,030 segments &
LiDAR \\

KAIST (RGB-T Ped.) &
Driving / pedestrian &
Vehicle &
Urban (day/night) &
RGB, TIR \\

FLIR ADAS &
Driving (thermal) &
Vehicle &
Road scenes &
RGB, TIR \\

\midrule
\multicolumn{6}{l}{\textit{Indoor robotics / 3D understanding}}\\
\midrule
ScanNet &
Indoor 3D &
Handheld RGB-D &
1,513 scans &
RGB, Depth \\

SUN RGB-D &
Indoor understanding &
RGB-D cameras &
Indoor rooms &
RGB, Depth \\

\midrule
\multicolumn{6}{l}{\textit{General-purpose vision / vision-language}}\\
\midrule
COCO &
General vision &
Web images &
Broad &
RGB \\

Open Images &
General vision &
Web images &
Broad &
RGB \\

\bottomrule
\end{tabularx}

\vspace{2mm}
\footnotesize
\noindent\textbf{Abbrev.} TIR: thermal infrared; Depth: depth point cloud; Radar: radar point cloud; Temp: temperature; Hum: humidity.
\end{table*}

Industrial vision datasets have primarily focused on quality inspection and defect recognition (as shown in Table \ref{tab:dataset_comparison}). Representative benchmarks include MVTec AD \cite{bergmann2019mvtec}, VisA \cite{zou2022spot}, Real IAD \cite{wang2024real}, Real IAD D3 \cite{zhu2025real}, MVTec 3D AD\cite{bergmann2021mvtec}, and Real3D AD\cite{liu2023real3d}. Data are typically collected on real production lines or fixed workstations, with clearly defined anomaly categories and pixel-level or precise defect annotations. Such benchmarks have significantly advanced research in anomaly detection and localization.

However, three major limitations are commonly observed. First, most data are captured at close range in static environments with controlled backgrounds and minimal disturbances. In practical inspection scenarios, environmental complexity is substantially higher, involving strong illumination variations, occlusion, specular reflection, sensor noise, and background clutter, which are insufficiently represented. Second, sensing modalities are usually limited to RGB or RGB combined with three-dimensional data. Additional modalities such as audio, radar point clouds, gas detection, and environmental parameters including temperature and humidity are rarely provided, despite their relevance to safety assessment. Third, annotations primarily focus on defects, while scene-level semantic descriptions and safety-related labels are generally missing. The absence of such information limits the evaluation of multimodal safety reasoning \cite{behzad2025robomnist}.
Further datasets have been introduced in autonomous driving and robotics, including KITTI \cite{geiger2013vision}, nuScenes\cite{caesar2020nuscenes}, and the Waymo Open Dataset\cite{sun2020scalability}. Visible and thermal datasets such as KAIST \cite{choi2018kaist} and FLIR ADAS support sensor synchronization, calibration, and large-scale three-dimensional perception evaluation. However, the semantic systems adopted in these datasets are designed for road environments, emphasizing vehicles, pedestrians, and traffic signs. In contrast, industrial inspection involves distinct semantics, including equipment categories, inspection site structures, and safety factors.

Several indoor RGB-D datasets have also been released, such as ScanNet \cite{dai2017scannet}, SUN RGB-D \cite{song2015sun}, and AVD\cite{ammirato2017dataset}. These datasets support dense geometric and semantic modeling, but do not provide industrial semantics or safety annotations for inspection. General-purpose vision and vision-language datasets, including COCO \cite{lin2014microsoft}, Open Images \cite{kuznetsova2020open}, and Visual Genome \cite{krishna2017visual}, offer large-scale object annotations and language supervision. However, the scenes are primarily drawn from daily life. Labels for industrial equipment, synchronized multi-sensor data from inspection robots, and safety-level annotations are not included.

\begin{figure*}[!htbp]
  \centering
      \includegraphics[width=0.97\textwidth]{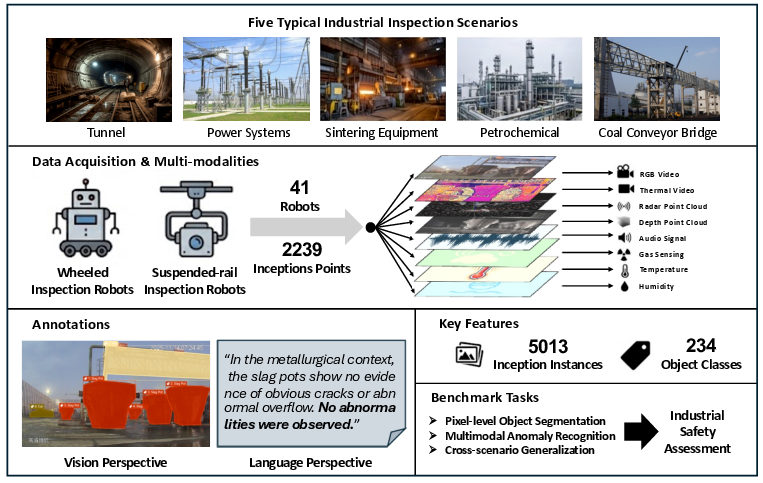}
  \caption{Overview of InspecSafe-V1 Benchmark Dataset.}
\label{fig_1}
\end{figure*}

In this context, the InspecSafe-V1 dataset is constructed and introduced in this paper. A structural overview is provided in Fig. \ref{fig_1}. The goal is to convert the dynamic complexity of real-world industrial inspection into a structured benchmark. InspecSafe-V1 is built entirely from raw data collected by frontline inspection robots and includes two types of platforms: wheeled and rail-mounted. Data acquisition covers five representative industrial environments, namely tunnels, power facilities, sintering equipment, oil and gas plants, and coal transfer trestles. A total of 41 robots and 2,239 inspection points are included. The dataset inherently captures realistic disturbances, including severe illumination variation, occlusion, specular reflection, sensor noise, and background clutter. These characteristics help reduce the distributional gap between idealized or simulated data and actual deployment scenarios. In terms of scale and annotation, InspecSafe-V1 includes 5,013 annotated instances with both visual and semantic labels, covering 234 key industrial inspection object categories. The dataset is expected to provide a valuable foundation for the development and evaluation of large multimodal models in industrial inspection.


\section*{Methods}\label{sec2}

\paragraph*{Setup}
InspecSafe-V1 is constructed based on real-world industrial inspection tasks. Data collection relies on two types of on-site deployed robotic platforms: wheeled inspection robots and rail-mounted inspection robots. The two platforms exhibit complementary characteristics in terms of mobility, viewpoint elevation, and spatial accessibility. The wheeled platform covers ground-level areas and provides close-range views of equipment, while the rail-mounted platform operates along fixed tracks, offering the ability to bypass obstacles and perform long-distance continuous inspections. The dual-platform configuration enables more comprehensive coverage of representative objects and safety elements within industrial environments. The collected data include key equipment (e.g., instruments, valves), basic components (e.g., screws), infrastructure elements (e.g., stairways, fire protection devices), and safety-related anomalies (e.g., smoke, unauthorized personnel access). The resulting object distribution and scene composition are closely aligned with real-world inspection processes.

\begin{figure*}[!htbp]
  \centering
      \includegraphics[width=0.97\textwidth]{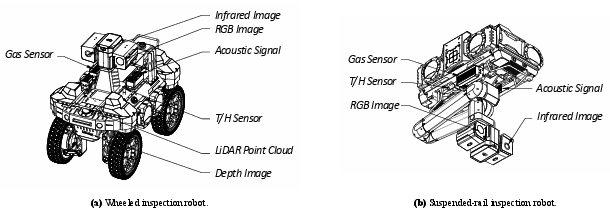}
  \caption{Illustration of the key sensor module configurations for wheeled and rail-mounted inspection platforms.}
    \label{fig:robot_platforms}
\end{figure*}


As shown in Fig. \ref{fig:robot_platforms}, the inspection platforms are equipped with a multimodal perception system designed for complex industrial environments. The core modules for geometric perception and localization include forward-facing RGB and depth cameras, as well as 3D LiDAR sensors. These components capture visual appearance, spatial geometry, and platform pose information, providing the foundation for temporal synchronization and spatial alignment of multimodal data.
Building on this configuration, the platforms further integrate various auxiliary sensors to enhance the perception of potential safety hazards. Thermal infrared cameras are typically co-located with RGB cameras on gimbals or front-mounted brackets to capture surface temperature distributions of equipment, supporting the detection of overheating and thermal anomalies. Millimeter-wave radar complements LiDAR by improving robustness in adverse conditions such as dust, low light, or partial occlusion, enabling reliable perception of structural elements and moving objects. Acoustic sensors are usually installed on the upper frame or internal structures to record machine noise and abnormal acoustic signals. Gas sensors are configured in a task-specific and scenario-dependent manner, with customizable gas types, measurement ranges, and sampling frequencies. For instance, combustible or toxic gas detection is prioritized in oil and gas facilities, whereas environmental safety monitoring is emphasized in tunnels and enclosed spaces.
Due to significant variations in process workflows and safety regulations across inspection scenarios, gas sensors may differ in model and configuration between platforms. Detailed specifications for each inspection scene are documented within the dataset.

For the specific hardware configuration, the RGB camera is equipped with a 1/2.8-inch CMOS sensor, offering a field of view (FoV) ranging from 55.8° to 2.3° horizontally, 31.9° to 1.3° vertically, and 63.7° diagonally. The frame rate is 25 FPS. The thermal infrared camera employs an uncooled detector with horizontal and vertical FoVs of 53.7° and 39.7°, respectively, and operates at the same frame rate of 25 FPS.
Depth perception is provided by the TM265-E1 depth camera from Orbbec, with an effective measurement range from 0.05 m to 5 m. Three-dimensional point cloud data are acquired using the MID360 LiDAR from LeiShen, with a maximum detection range of up to 40 m under 10\% reflectivity and 70 m under 80\% reflectivity.




It is worth noting that the modality configurations differ between the two types of inspection platforms. Influenced by structural design, payload capacity, and specific inspection requirements, the wheeled and rail-mounted platforms vary in both sensor setups and accessible sensing modalities. The wheeled platform typically integrates RGB and thermal imaging, depth sensing, 3D LiDAR, acoustic sensing, and environmental perception modalities such as gas detection and temperature/humidity monitoring. This configuration enables the platform to capture both close-range equipment details and large-scale geometric structures. In contrast, the rail-mounted platform is constrained by limited installation space and motion patterns, and thus primarily focuses on forward-facing visual and thermal imaging, acoustic sensing, and, in certain scenarios, selected environmental sensors. The configuration is optimized for continuous monitoring of key infrastructure and equipment along the track.
Although differences exist in sensor types and modality combinations, all inspection instances in the dataset consistently provide RGB imagery along with corresponding language-level annotations. For each instance, the RGB data are annotated with pixel-level instance segmentation masks, while the language annotations include semantic scene descriptions and safety level labels. This ensures semantic consistency across data collected by different platforms.

At the data level, InspecSafe-V1 adopts the inspection instance as the fundamental organizational unit. Multimodal data are structured in a unified format and aligned across modalities via timestamp synchronization and coordinate transformation. This alignment guarantees consistency in both structure and semantics across platforms and inspection environments, allowing all data to be integrated into downstream modeling, training, and evaluation pipelines without disruption from platform-specific differences.

\paragraph*{Data Acquisition}

Data acquisition is organized using inspection points as the fundamental unit. Upon arrival at each predefined point, the inspection robot performs synchronized multimodal recording within a short time window, enabling unified packaging, indexing, and annotation of visual, point cloud, acoustic, and environmental sensing data corresponding to the same spatial location. Each inspection point refers to a planned stop along the inspection route, where the robot is instructed to pause during execution. These points are typically situated at key equipment locations or safety-critical areas and serve as standard observation positions. During the stop, the robot remains stationary for a short duration to facilitate synchronized multimodal data collection, ensuring spatial and temporal consistency across modalities. This consistency is critical for subsequent data organization and annotation. To protect personal privacy, some videos containing human subjects were processed before public release, with all visible human faces anonymized by mosaic masking.

In practical deployment, the dwell time at each inspection point is approximately 10 seconds to 15 seconds. During this period, RGB video, thermal infrared video, and audio are recorded continuously, resulting in point-level audiovisual data. Meanwhile, 3D point cloud data are captured over a span of approximately 3 seconds to obtain stable geometric representations. As a result, each inspection point is associated with a 10–15 second RGB video segment, temporally aligned with the corresponding thermal, acoustic, and point cloud data, providing a comprehensive multimodal description of the location.

Data collection spans multiple months and includes both daytime and nighttime sessions to reflect natural temporal variations present in real industrial inspection environments. For selected key points, repeated acquisitions are performed to better characterize temporal dynamics and environmental disturbances. Specifically, each 24-hour day is divided into two time segments: 00:00–12:00 and 12:00–24:00. Data acquisition is triggered only during the first inspection task executed within each time segment, resulting in a maximum of two recordings per point per day. This strategy captures diurnal variation and operational changes, such as shifts in lighting and process rhythms.

\begin{figure*}[!htbp]
  \centering
      \includegraphics[width=0.97\textwidth]{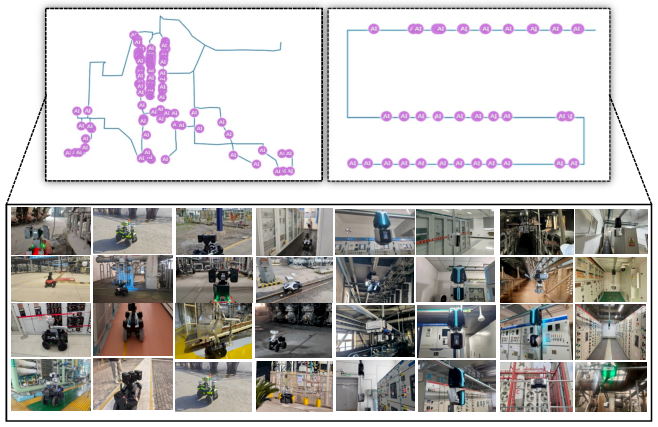}
  \caption{Illustration of the data acquisition process and sampling trajectories.}
\label{fig_3}
\end{figure*}

Illustrative examples of the data collection environment are shown in Fig. \ref{fig_3}. The selection and deployment strategy of key inspection points are tailored to specific task scenarios.

At each inspection point, RGB video is stored in MP4 format with a resolution of either 2560×1440 or 1920×1080 pixels, and a recording duration of 10 to 15 seconds per clip. Thermal video is also saved in MP4 format, with a resolution of either 1280×1024 or 640×480 pixels, and the same recording duration of 10 to 15 seconds. Environmental audio is stored in WAV format with a sampling rate of 8000 Hz, two channels, and a recording duration of 10 to 15 seconds per clip.
Point cloud data are recorded in ROS bag format, with a sampling duration of 3 seconds per point at a rate of 15 Hz. Environmental sensing data, including gas concentration, temperature, and humidity, are stored in plain text files and temporally aligned with the inspection point timestamps. All modalities are associated through a unified inspection point identifier and synchronized timestamps. Metadata entries record the start and end times of each modality and indicate their availability, thereby supporting cross-modal alignment, unified indexing, and downstream annotation processes.
Point cloud acquisition is performed either at the beginning of the inspection point window or within a predefined sub-window. The captured point cloud data are aligned with RGB video, thermal video, audio, and environmental sensor data based on the global timestamp system. A summary of the synchronized modalities, sensor types, and data formats used at each inspection point in InspecSafe-V1 is provided in Table \ref{tab:modality_storage}.

\begin{table}[!htbp]
\centering
\caption{Multimodal acquisition configuration and data storage formats.}
\label{tab:modality_storage}
\small
\renewcommand{\arraystretch}{1.2}
\begin{tabular}{l l l l p{4.7cm}}
\toprule
\textbf{Modality} & \textbf{Sensor Type} & \textbf{File Format} & \textbf{Extension} & \textbf{Remarks} \\
\midrule
RGB Video & Camera & MP4 & .mp4 & 1920×1080 / 2560×1440; 10–15 s \\
Thermal Video & Thermal Camera & MP4 & .mp4 & 1280×1024 / 640×480; 10–15 s \\
Point Cloud & Depth Camera, LiDAR & ROS Bag & .bag & 15 Hz; 3 s \\
Audio & Microphone Array & WAV & .wav & 8000 Hz; stereo; 10–15 s \\
Others & Gas, Temp, Humidity Sensors & TXT & .txt & —— \\
\bottomrule
\end{tabular}
\end{table}

\paragraph*{Pipeline}

Unlike frame-by-frame annotation, InspecSafe-V1 employs a fixed-step frame sampling strategy to reduce annotation redundancy, considering the strong visual and spatial consistency between adjacent short-term frames. Specifically, one frame is extracted every five frames from video sequences to serve as a candidate for annotation. To further suppress redundancy caused by repetitive viewpoints and near-static scenes, a similarity-based frame filtering mechanism is introduced. By comparing deep visual features between adjacent candidates, highly similar frames are discarded, thereby increasing the diversity of objects and expanding coverage under the same annotation budget. Sampled and filtered images serve as inputs for visual annotation, and are linked to thermal video, audio, point cloud, and environmental sensing data via inspection point identifiers and timestamps. 

\begin{figure*}[!htbp]
  \centering
      \includegraphics[width=0.97\textwidth]{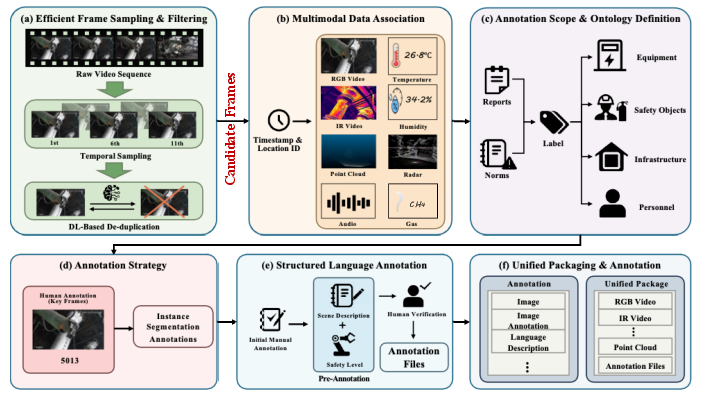}
  \caption{Overview of the InspecSafe-V1 dataset construction pipeline.}
\label{fig_4}
\end{figure*}

The pipeline is summarized as Fig. \ref{fig_4}. The {\it reports} refer to inspection reports, defect records, and related documentation collected during routine inspection and maintenance procedures by both robots and human personnel, while the {\it norms} refer to industry standards, technical specifications, and operational procedures used as the basis for determining safety risks and compliance. In total, five annotators participated in the fine-grained image segmentation annotation, and two annotators were responsible for semantic annotation.

During the construction and execution of the annotation protocol, labeling scope is defined based on objects of interest in industrial inspection tasks. A structured and extensible label taxonomy is established. For industrial equipment, standardized names are extracted from inspection reports and asset lists to construct the label set. To support safety understanding, semantically salient and safety-critical objects are also included, such as personnel, firefighting facilities, staircases, and personal protective equipment like helmets.
Due to the frequent co-occurrence of multiple objects in real inspection scenes, each annotated frame typically contains several labeled instances. Instance-level segmentation is performed using polygons to achieve pixel-level precision, accurately outlining object boundaries while minimizing background interference. This approach ensures high-quality supervision for downstream tasks such as segmentation, object-level reasoning, and structured scene modeling.

In addition to visual labels, structured language-level annotations are provided to support multi-layer modeling from perception to safety reasoning. Language annotations include scene descriptions and safety level assessments. Scene descriptions summarize environmental context, key objects, and visible events, while safety levels are assigned according to industrial inspection safety standards and dataset-specific labeling guidelines. The criteria for assigning safety levels are detailed in Table \ref{tab:safety_levels_transposed}. Notably, if no safety factors are present, assign the safety level as {Level IV}.

\begin{table*}[!htbp]
\centering
\caption{Key criteria for safety level classification across different industrial scenarios.}
\label{tab:safety_levels_transposed}
\small
\renewcommand{\arraystretch}{1.25}
\setlength{\tabcolsep}{5pt}
\begin{tabular}{l p{4.2cm} p{4.2cm} p{4.2cm}}
\toprule
\textbf{Scenario} & \textbf{Level I} & \textbf{Level II} & \textbf{Level III} \\
\midrule
Oil \& Gas / Chemical &
Open flame, smoke, no hard hat, no gloves, no mask, smoking, personnel collapse, oil leakage &
Water pooling, using mobile phone &
Foreign objects \\
\midrule
Coal Conveyor Gallery &
Open flame, smoke, no hard hat, smoking, personnel collapse &
Using mobile phone, no gloves, no mask, foreign objects (plastic bags, bottles, foam, metal, paper) &
Water pooling \\
\midrule
Tunnel &
Open flame, smoke, non-motorized vehicles in fast lane, wood, metal, personnel collapse &
Foam, plastic bags, plastic bottles, no hard hat, cabinet door open &
Water pooling, oil accumulation, using mobile phone, no gloves, no mask, smoking \\
\midrule
Power &
Open flame, smoke, smoking, personnel collapse &
Water pooling, oil accumulation, using mobile phone, foreign objects (plastic bags, bottles, foam, metal, paper), no hard hat, no gloves, cabinet door open &
No mask \\
\midrule
Metallurgy &
Open flame, smoke, no hard hat, personnel collapse &
No gloves, no mask, smoking &
Water pooling, oil accumulation, using mobile phone, foreign objects (plastic bags, bottles, foam, metal, paper) \\
\bottomrule
\end{tabular}
\end{table*}

All data are uniformly packaged and indexed. Each modality is linked through inspection point identifiers and timestamps. Annotation files maintain traceable relationships with raw multimodal data via metadata records, which include scene ID, robot ID, inspection point ID, timestamps, modality availability, and file paths. Such organization ensures data traceability, reproducibility, and compatibility with benchmark task definitions and evaluation protocols.

\section*{Data Records}\label{sec3}

The dataset is archived in the Zenodo repository \cite{liu2026inspecsafe} and is additionally accessible through Hugging Face repository at
\url{https://huggingface.co/datasets/Tetrabot2026/InspecSafe-V1}. The repository contains two subsets, \textit{train} and \textit{test}. Each subset includes three folders: \textit{Annotations}, \textit{Other\_modalities}, and \textit{Parameters}. The same folder organization is used in both subsets. A unified README.md file is provided at the root level to describe the dataset's contents and usage instructions.

The file structure of the dataset is organized using the Inspection Point as the fundamental unit. Each inspection point corresponds to a folder (e.g., \texttt{tunnel-Level04-SuspendedRail-000539}), which aggregates the annotations and multimodal records for that location. A single inspection point may contain multiple Inspection Instances (e.g., instances taken at different times or multiple frames from the same point). In the \textit{Annotations} folder, these instances are distinguished by numerical suffixes appended to the filenames (e.g., {-001}, {-002}, {-003}), with each suffix corresponding to a specific set of annotation files. The corresponding \textit{Other\_modalities} directory shares the identical point identifier, ensuring that the multimodal and sensory records can be accurately linked to one or more inspection instances at that location. A detailed illustration of the full directory structure is provided in Supplementary S1.

\paragraph{Annotations.}
The \textit{Annotations} folder is divided into \textit{Normal\_data} and \textit{Anomaly\_data}, which contain normal and abnormal inspection instances, respectively. Each inspection sample is stored in an individual directory named according to the inspection object, location, and sample identifier, such as \texttt{coal\_conveyor-Level04-SuspendedRail-000560}. Each sample directory contains the annotated image frame and its corresponding annotation files, including a \textit{.jpg} image file, a \textit{.json} annotation file, and a \textit{.txt} textual description. The \textit{.json} file follows a LabelMe-style format. Annotated instances are stored in the \textit{shapes} field. For each instance, \textit{label} denotes the object or anomaly category, \textit{points} records the polygon vertices, and \textit{shape\_type} is set to \textit{polygon}. The image resolution is provided by \textit{imageHeight} and \textit{imageWidth}. The field \textit{imageData} stores the embedded image content, while \textit{imagePath} records the original frame name. The \textit{.txt} file provides a plain-text description of the corresponding inspection point.

\paragraph{Other\_modalities.}
The \textit{Other\_modalities} folder contains multimodal records associated with each inspection sample. Each sample directory is named using the same inspection identifier as in \textit{Annotations}, which allows the annotation files and multimodal records to be matched. For each sample, the \textit{-visible.mp4} file contains the RGB video, the \textit{-infrared.mp4} file contains the infrared video, the \textit{-audio.wav} file contains the audio record, and the \textit{-point\_cloud.bag} file contains the point-cloud data. The \textit{-sensor.txt} file stores structured sensor and environmental status information in JSON format. For example, the \textit{env} field contains a list of sensor records, where \textit{name}, \textit{unit}, and \textit{value} denote the sensor name, measurement unit, and recorded value, respectively.

\paragraph{Parameters.}
The \textit{Parameters} folder provides device parameter specifications as JSON files, including \textit{RGB\_Camera\_params.json}, \textit{IR\_Camera\_params.json}, \textit{Depth\_Camera\_params.json}, and \textit{LiDAR\_params.json}. These files record modality-specific device information and acquisition parameters, such as device identifiers, resolution, intrinsic parameters, extrinsic parameters, and other configuration fields where applicable.

\color{black}

\section*{Data Overview}\label{secdata}

\begin{figure}[t]
  \centering
      \includegraphics[width=0.97\textwidth]{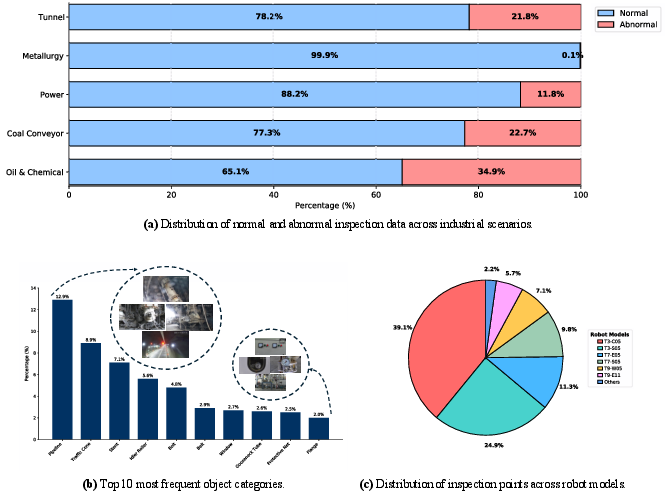}
  \caption{Dataset statistics of InspecSafe-V1 across scenarios, object categories, and robot platforms.}
\label{fig:datafields}
\end{figure}

Based on robot model statistics, as shown in Fig. \ref{fig:datafields}, the distribution of inspection points across different platforms is notably imbalanced. The T3 C05 and T3 S05 models, both rail-mounted platforms, account for 39.1\% and 24.9\% of the inspection points respectively, contributing a combined total of 64.0\%. T7 E05 and T7 S05 account for 11.3\% and 9.8\%, while T9 W05 and T9 E11 contribute 7.1\% and 5.7\%. The remaining models collectively account for only 2.2\%. This distribution reflects real-world deployment conditions where mainstream platforms contribute the majority of the data and long-tail platforms enhance diversity. Rail-mounted platforms dominate in terms of data volume, providing a practical basis for research on cross-platform generalization, platform bias analysis, and evaluation under heterogeneous modality settings.

All inspection instances in InspecSafe-V1 are uniformly annotated with RGB keyframes, pixel-level object masks, scene descriptions, and safety-level labels. A total of 234 object categories are defined in the RGB modality, showing a significant long-tail distribution. The most frequent object categories include Pipeline (12.9\%), Traffic Cone (8.9\%), Stent (7.1\%), Idler Roller (5.6\%), and Bolt (4.8\%), which together account for 39.3\%. Including Belt (2.9\%), Window (2.7\%), Gooseneck Tube (2.6\%), Protective Net (2.5\%), and Flange (2.0\%), the top ten categories account for 52.0\%. This distribution is consistent with real-world appearance frequencies and provides a natural and challenging setting for learning under class imbalance.

Across industrial domains, the dataset includes tunnel, power, metallurgy, coal transportation, and oil and gas scenarios. The normal and abnormal sample ratios vary considerably by domain. In tunnel scenes, 78.2\% of instances are normal and 21.8\% are abnormal. In power scenes, the ratio is 88.2\% to 11.8\%. In coal transportation, it is 77.3\% to 22.7\%. In oil and gas, abnormal instances account for 34.9\% and normal instances for 65.1\%. In metallurgy, nearly all instances are normal, with abnormal cases accounting for only 0.1\%. These statistics reflect substantial differences in anomaly occurrence across domains.


\section*{Technical Validation}\label{sec4}

\paragraph*{Data Quality Assessment}

During data organization and annotation, RGB video serves as the primary visual modality. For each inspection point, representative frames are selected from the captured video based on a predefined keyframe extraction strategy. These frames are then manually annotated with fine-grained labels. 
{These selected frames preserve challenging real-world factors commonly encountered in industrial inspection scenarios, such as reflections, illumination and shadow interference, and sensor noise (e.g., camera blur). It is also indicated that these artifact-related factors account for a noticeable proportion of model misrecognitions across different models.}
{All visual annotations are conducted using the Label Studio platform, an open-source annotation tool\cite{labelstudio}.}

{To ensure the reliability of the pixel-level annotations, all annotated images are subjected to a two-round independent verification procedure. In each round, 5\% of the annotated instances are randomly selected for quality inspection. If any object in a sampled image is found to be mislabeled, inaccurately annotated, or missing, the image is recorded as erroneous. Only annotations achieving an accuracy higher than 95\% in both verification rounds are retained after quality control and incorporated into the final dataset.}

In addition to pixel-level annotation, semantic annotations also undergo a structured quality assessment process. Ten percent of the semantic annotation results are randomly selected and reviewed through the same two-round independent verification mechanism. The accuracy of semantic annotations is evaluated based on the following criteria:

\begin{enumerate}
\item[1)] Whether the scene description correctly captures the semantic context of the inspection point.

\item[2)] Whether the assigned safety level accurately reflects the observed conditions and potential safety threats present in the scene.

\item[3)] Whether key objects and potential hazard factors are completely and accurately described, with particular attention to safety-critical information.

\end{enumerate}

For safety level annotations, if multiple hazards of different safety threats appear in the same image, the overall safety level is determined by the most severe hazard present. All three types of semantic annotations must meet the consistency and accuracy requirements in the sampled data in order to pass the quality audit and be included in the final released dataset.

\paragraph*{Benchmark Evaluation}

The benchmark evaluation targets safety assessment in real-world industrial inspection scenarios. The primary goal is to quantitatively evaluate the semantic understanding capabilities of general-purpose \textit{vision-language models} (VLMs) under complex environmental conditions \cite{shinde2025survey,bordes2024introduction,dai2023instructblip}. The dataset is organized based on inspection points. To minimize potential information leakage and sampling bias caused by high visual similarity between adjacent frames within the same location, both the training and testing sets are constructed using uniform intra-point sampling. Keyframes are selected at fixed intervals from RGB video sequences, forming representative samples for evaluation. The final split includes 3,763 training frames (3,014 normal and 749 abnormal) and 1,250 testing frames (999 normal and 251 abnormal). It preserves the natural class imbalance while ensuring broader coverage of diverse scenes, thereby enhancing the representativeness of evaluation results.

Each RGB frame $x_i$, together with a standardized prompt template $p$, is input to the target model $f_\theta(\cdot)$, which is required to produce two outputs: a natural language scene description $\hat{s}_i$, and a discrete safety level prediction $\hat{y}_i$. The safety level is selected from a predefined label set, and the output is parsed into structured format using strict rule-based field parsing. Prediction correctness is determined by comparing $\hat{y}_i$ against the ground truth label $y_i$. {Referred to the literature\cite{zhou2022learning}, the utilized prompt for benchmark evaluation is reported in Supplementary S2.}
The accuracy of safety level prediction is computed as follows:
\begin{equation}
\mathrm{Acc} = \frac{1}{N} \sum_{i=1}^N \mathbf{1}\left(\hat{y}_i = y_i\right),
\end{equation}
where $N = 1250$ is the total number of test samples, and $\mathbf{1}(\cdot)$ denotes the indicator function.

Beyond label-level prediction, semantic consistency is evaluated by comparing the generated descriptions $\hat{s}_i$ with the annotated scene descriptions $s_i$. A fixed text encoder $g(\cdot)$, specifically BGE-M3 \cite{chen2024bge}, maps both texts into a shared embedding space. The semantic similarity for each sample is calculated using cosine similarity:
\begin{equation}
\operatorname{SemSim} = \frac{1}{N} \sum_{i=1}^N \operatorname{sim}_i = \frac{g(\hat{s}_i)^\top g(s_i)}{\|g(\hat{s}_i)\|_2 \cdot \|g(s_i)\|_2},
\end{equation}
which quantifies the alignment between model-generated descriptions and human annotations in terms of key objects, scene context, and safety-related semantics. It provides complementary insight beyond discrete classification accuracy.

\begin{figure*}[!htbp]
  \centering
      \includegraphics[width=0.90\textwidth]{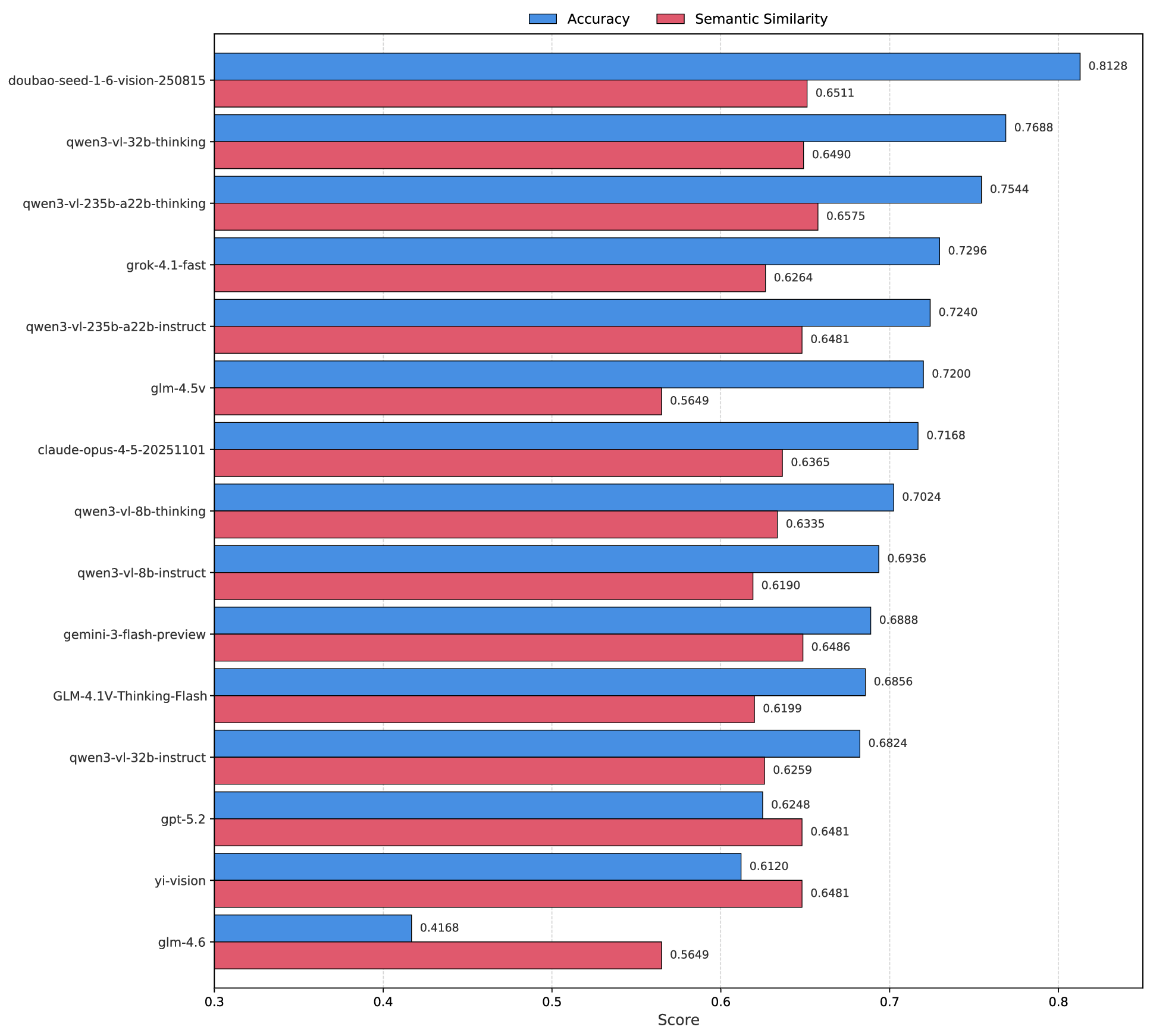}
  \caption{Benchmark results of different VLMs on InspecSafe-V1.}
\label{fig_6}
\end{figure*}

Evaluation results are shown in Fig. \ref{fig_6}. The findings reveal that accuracy and semantic similarity do not exhibit a strictly monotonic relationship with model size. Performance is more closely influenced by the alignment between perceptual robustness and reasoning capability. Some models with smaller parameter sizes outperform larger models, indicating that increased parameter count does not necessarily lead to more reliable safety judgment.

Reasoning-oriented models consistently achieve better performance compared to their instruction-only counterparts. Within the Qwen3-VL family, the reasoning-enhanced variant achieves approximately eight percentage points higher accuracy than the instruction-tuned model, while also reducing the number of false positives. This demonstrates that incorporating intermediate reasoning steps improves evidence integration and prediction consistency in visually complex industrial scenes.

{Confusion matrices for each model, summarized in Supplementary S3, further highlight differences in strategies.} Variations in false positive counts indicate differing levels of conservativeness across models. Most false positives originate not from actual hazards but from visual ambiguity caused by harsh lighting, strong reflections, backlighting, high-contrast shadows, or particle scattering. These conditions frequently appear in real industrial environments and present a considerable challenge. Some models exhibit strong resilience, such as Doubao-SEED-1-6-Vision-250815. Others, including GPT-5.2, apply more conservative safety thresholds, meaning more than thirty percent of normal content is misclassified, which could lead to excessive alarm load and increased verification costs during deployment.

In addition to false positives, two representative error patterns are observed. The first involves cascading failures triggered by incorrect scene recognition. Some models, such as GLM-4.6, misclassify coal-handling environments as oil and chemical facilities. Since safety standards are highly context-dependent, incorrect scene classification disrupts the entire reasoning pipeline and leads to inaccurate safety assessments. The second error pattern involves the failure to detect clear violations, such as the use of mobile phones, absence of safety gloves, or missing helmets. These omissions are often associated with small object size, occlusion, or insufficient training coverage, exposing the limitations of general-purpose models in fine-grained behavior recognition tasks.

The benchmark evaluation highlights two core challenges for applying VLMs to industrial safety assessment: robust safety threat recognition under challenging visual conditions and context-aware reasoning for fine-grained violation detection. Models with integrated reasoning capabilities demonstrate clear performance advantages in both dimensions. At the same time, false positive rates reflect the balance between model sensitivity and operational feasibility, offering actionable guidance for model selection, prompt engineering, and threshold setting in real-world deployment.

\section*{Data Availability}\label{sec5}
The dataset is archived in the Zenodo repository~\cite{liu2026inspecsafe}. The dataset and its associated resources are also publicly accessible through Hugging Face repository at
\url{https://huggingface.co/datasets/Tetrabot2026/InspecSafe-V1}.
All data involving privacy- or safety-sensitive content have been anonymized or removed in accordance with
applicable regulations prior to release.
\color{black}

\section*{Usage Notes}\label{sec6}

Of note, the InspecSafe-V1 dataset is primarily designed for safety assessment tasks in industrial inspection scenarios. However, its utility extends broadly to a variety of research areas, including multimodal perception and representation learning, cross-modal fusion, vision-language joint modeling, cross-domain generalization, and domain adaptation. Each inspection instance in the dataset is paired with an RGB image, pixel-level object segmentation annotation, and a corresponding semantic textual description, providing a robust foundation for multimodal understanding and reasoning in real-world industrial environments. {The highly imbalanced distribution in some domains, such as metallurgy, should be considered when using the dataset for model evaluation. In such settings, evaluation protocols may need to account for class imbalance, false positive rates, and cross-domain generalization.}

{In terms of data formatting and organization, the image annotations are stored in JSON format and remain fully compatible with mainstream polygon annotation tools, such as X-AnyLabeling (\url{https://github.com/CVHub520/X-AnyLabeling}), which facilitates efficient visualization, annotation review, and subsequent extension.}
All multimodal data are organized at the inspection-instance level, with consistent naming conventions and timestamp alignment to support cross-modal synchronization. Researchers can flexibly select single-modality or multimodal subsets depending on their specific modeling and evaluation needs. While sensor configurations and available modalities may vary across platforms, every inspection instance uniformly includes RGB imagery along with corresponding pixel-level annotations and scene-level semantic labels, ensuring consistency and completeness in the semantic layer.

{It is worth noting that InspecSafe-V1 provides a systematic foundation for multimodal safety assessment in real-world industrial inspection scenarios. The current benchmark mainly represents scene safety through discrete semantic labels, which offer clear and practical supervision for engineering applications. Owing to real-world inspection and annotation constraints, the present version can be further expanded in terms of dataset scale, scene diversity, extreme operating conditions, rare hazard types, and long-duration sequential inspection data.}

\section*{Code Availability}\label{sec7}

The source code used in this study has been made publicly available at \url{https://github.com/liuzy0708/InspecSafe} and is also accessible through \url{https://huggingface.co/datasets/Tetrabot2026/InspecSafe-V1}.

\color{black}





\section*{Acknowledgements}
This work was supported in part by National Natural Science Foundation of China under grants 62525308, 624B2087, 62473223, and 52172323, in part by Beijing Natural Science Foundation under grant L241016. We also thank all research assistants who provided general support in participant recruiting and data collection. We additionally thank Jiali Chen, Jinbao Zhang and Chen Li for their contribution to data organization and curation.

\section*{Author Contributions}
Z. Liu and S. Liu led the methodology, software implementation, formal analysis, and writing of the original draft. J. Min, Z. Zhang, and Z. Meng contributed to data acquisition, data curation, and provision of resources. J. Cen provided methodological guidance and technical input. P. Han and S. Hu contributed to software development, experiment execution, validation, and data processing. X. He and D. Zhou supervised the study and managed project coordination.

\section*{Competing Interests}
The authors declare no competing interests.

\section*{Corresponding author}
Correspondence to Prof. Xiao He (email: hexiao@tsinghua.edu.cn).

\end{document}